\documentclass{article}

\usepackage{float}
\usepackage{microtype}
\usepackage{graphicx}
\usepackage{subfigure}
\usepackage{booktabs} 
\usepackage{amsmath}
\usepackage{amssymb}
\usepackage{bbm}
\usepackage{hyperref}
\usepackage{xspace}

\usepackage[normalem]{ulem}
\usepackage[shortlabels]{enumitem}
                \setlist[enumerate, 1]{1\textsuperscript{o}}
\usepackage[usenames,dvipsnames]{xcolor}
\usepackage{tikz}
\usetikzlibrary{shapes, calc}
\usepackage{verbatim}

\DeclareRobustCommand{\eg}{e.g.,\@\xspace}
\DeclareRobustCommand{\ie}{i.e.,\@\xspace}

\DeclareRobustCommand{\wrt}{w.r.t.\@\xspace}

\newcommand{\bnp}{TI\xspace}

\usepackage[accepted]{icml2020}

\icmltitlerunning{Learning Adaptive Exploration Strategies in Dynamic Environments}

\begin{document}

\twocolumn[
\icmltitle{
Learning Adaptive Exploration Strategies in Dynamic Environments\\ Through Informed Policy Regularization
}
\icmlsetsymbol{equal}{*}

\begin{icmlauthorlist}
\icmlauthor{Pierre-Alexandre Kamienny}{fb}
\icmlauthor{Matteo Pirotta}{fb}
\icmlauthor{Alessandro Lazaric}{fb}
\icmlauthor{Thibault Lavril}{fb}
\icmlauthor{Nicolas Usunier}{fb}
\icmlauthor{Ludovic Denoyer}{fb}
\end{icmlauthorlist}

\icmlaffiliation{fb}{Facebook AI Research, Paris, France}

\icmlcorrespondingauthor{Ludovic Denoyer}{denoyer@fb.com}

\vskip 0.3in
]
\printAffiliationsAndNotice{}
\begin{abstract}
We study the problem of learning exploration-exploitation strategies that effectively adapt to dynamic environments, where the task may change over time. While RNN-based policies could in principle represent such strategies, in practice their training time is prohibitive and the learning process often converges to poor solutions. In this paper, we consider the case where the agent has access to a description of the task (e.g., a task id or task parameters) at training time, but not at test time. We propose a novel algorithm that regularizes the training of an RNN-based policy using informed policies trained to maximize the reward in each task. This dramatically reduces the sample complexity of training RNN-based policies, without losing their representational power. As a result, our method learns exploration strategies that efficiently balance between gathering information about the unknown and changing task and maximizing the reward over time. We test the performance of our algorithm in a variety of environments where tasks may vary within each episode.

\end{abstract}

\section{Introduction}
\label{introduction}

Deep Reinforcement Learning (DRL) has been used to successfully train agents on a range of challenging environments such as Atari games~\citep{atari, ALE, rainbow} or continuous control~\citep{sim2real, ppo}. 
Nonetheless, in these problems, RL agents perform exploration strategies to discover the environment and implement algorithms to learn a policy that is tailored to solving a \emph{single task}. Whenever the task changes, RL agents generalize poorly and the whole process of exploration and learning restarts from scratch. 
On the other hand, we expect an intelligent agent to fully master a \textit{problem} when it is able to generalize from a few instances (tasks) and learn how to achieve the objective of the problem under many variations of the environment. For instance, children \textit{know} how to ride a bike (i.e., the problem) when they can reach their destination irrespective of the specific bike they are riding, which requires to adapt to the weight of the bike, the friction of the brakes and tires, and the road conditions (i.e., the tasks). 

How to enable agents to generalize across tasks has been studied under the frameworks of 
\textit{Multi-task Reinforcement Learning}~\citep[\eg][]{multitask_RL, distral}, \textit{Transfer Learning}~\citep[\eg][]{transfer_rl, lazaric2012transfer} and \textit{Meta-Reinforcement Learning}~\citep{MAML,Hausman2018LatentSpace,Rakelly2019offpolicy,task_inference}. 
These works fall into two categories. \emph{Learning to learn} approaches aim at speeding up learning on new tasks, by pre-training feature extractors or learning good initializations of policy weights \citep{MAML_analysis}. In contrast, we study in this paper the \emph{online adaptation} setting where a single policy is trained for a fixed family of tasks. When facing a new task, the policy must then balance \emph{exploration}, to reduce the uncertainty about the current task, and \emph{exploitation} to maximize the cumulative reward of the task.

The online adaptation setting is a special case of a partially observed markov decision problem, where the unobserved variables are the descriptors of the current task. It is thus possible to rely on recurrent neural networks (RNNs)~\citep{Bakker2001LSTM, control_rnn, Duan2016RL2}, since they can theoretically represent optimal policies in POMDPs if given enough capacity. Unfortunately, the training of RNN policies has often prohibitive sample complexity and it may converge to suboptimal local minima.

To overcome this drawback, efficient online adaptation methods leverage the knowledge of the task at train time. The main approach is to pair an exploration strategy with the training of \textit{informed policies},  \ie policies taking the description of the current task as input. 
\emph{Probe-then-Exploit} (PTE) algorithms~\citep[\eg][]{zhou2019probing} operate in two stages. They first rely on an exploration policy to identify the task. Then, they commit to the identified task by playing the associated informed policy. 
\emph{Thompson Sampling} (TS) approaches~\citep{thompson1933likelihood,Osband2016bootstrappeddqn,Osband2019deepexpl} maintain a distribution over plausible tasks and play the informed policy of a task sampled from the posterior following a predefined schedule.

PTE and TS are expected to be sample-efficient because learning informed policies is easier than RNN policies: since informed policies know the current task, the problem is fully observable and they can be learnt with efficient algorithms for MDPs. However, as we discuss in Section \ref{sec:related_work}, PTE and TS cannot represent effective exploration/exploitation policies in many environments. 
This limitation is even more severe in \textit{non-stationary environments}, where the task changes within each episode. 
In this case, the exploration strategy must also adapt to how tasks evolve over time.

Recently, \citet{task_inference} proposed an alternative approach, \emph{Task Inference} (\bnp), which trains a full RNN policy with the prediction of the current task as an auxiliary loss. \bnp avoids the suboptimality of PTE/TS since it does not constrain the structure of the exploration/exploitation policy. However, in \bnp, the task descriptors are used as targets and not as inputs, so \bnp does not leverage the faster learning of informed policies. Moreover, the behavior of \bnp in non-stationary environments has not been studied. 

In this paper, we introduce IMPORT (\textit{InforMed POlicy RegularizaTion}), a novel policy architecture for efficient online adaptation that combines the rich expressivity of RNNs with the efficient learning of informed policy. At training time, a shared policy head receives as input the current observation, together with either a (learned) embedding of the current task, or the hidden state of an RNN such that two policies are learned simultaneously: the informed policy and the RNN policy. At test time, the hidden state of the RNN replaces the task embedding, and the agent can act without having access to the current task. 

We evaluate IMPORT against the main approaches to online adaptation on a suite of different environments with different characteristics, from challenging exploration problems with sparse rewards to non-stationary control problems. We confirm that TS suffers from its limited expressivity when non-trivial probing policies are required, and show that the policy regularization of IMPORT significantly speeds up learning compared to \bnp. Moreover, the learnt task embeddings of IMPORT make it robust to irrelevant or minimally informative task descriptors, while TI performances degrade significantly when task descriptors contain irrelevant variables or are only minimally informative.
\section{Setting}
\label{sec:setting}
Let $\mathcal{M}$ be the space of possible tasks. Each $\mu \in \mathcal{M}$ is associated to an episodic $\mu$-MDP $M_\mu = (\mathcal{S}, \mathcal{A}, p^\mu, r^\mu, \gamma)$ whose dynamics $p^\mu$ and rewards $r^\mu$ are task dependent, while state and action spaces are shared across tasks and $\gamma$ is the discount factor. The descriptor $\mu$ can be a simple id ($\mu \in \mathbb{N}$) or a set of parameters ($\mu \in \mathbb{R}^d$). 

When the reward function and the transition probabilities are unknown, RL agents need to devise a strategy that balances exploration to gather information about the system and exploitation to maximize the cumulative reward. Such a strategy can be defined as the solution of a partially observable MDP (POMDP), where the hidden variable is the descriptor $\mu$ of the MDP. Given a trajectory $\tau_{t} = (s_1, a_1, r_1, \ldots, s_{t-1}, a_{t-1}, r_{t-1}, s_t)$, a POMDP policy $\pi(a_t| \tau_t)$ maps the trajectory to actions. In particular, the optimal policy in a POMDP is a history-dependent policy that uses $\tau_t$ to construct a belief state $b_t$, which describes the uncertainty about the task at hand, and then maps it to the action that maximizes the expected sum of rewards~\citep[\eg][]{Kaelbling1998POMDP}. In this case, maximizing the rewards may require taking explorative actions that improve the belief state enough so that future actions can be more effective in collecting reward. 

At training time, we assume the agent has unrestricted access to the descriptor $\mu$ of the tasks it interacts with. In particular, we consider the challenging case of dynamic (i.e., non-stationary) environments, where the task may change over time according to a fixed or random schedule, $\mu_t$ being the value of $\mu$ at time $t$. Let $q(\mu | \tau_t, \{\mu_i\}_{i=1}^{t-1})$ to be a history-dependent task distribution, then at each step $t$ a new task $\mu_t$ is drawn from $q$.\footnote{Notice that the definition of $q$ is rich enough so that it can represent cases such as stationary, piece-wise stationary, and shifts with limited deviations.} Leveraging the information gathered at training time, we expect the agent to learn an exploration strategy that is better suited for tasks in $\mathcal{M}$ and $q$. More formally, after $T$ steps of training, the agent returns a policy $\pi(a_t| \tau_t)$ that is evaluated according to its average performance across tasks in $\mathcal{M}$ generated from $q$, i.e., 
\begin{align}\label{eq:test.performance}
\mathbb{E} \bigg[ \sum_{t=1}^{|\tau|} \gamma^{t-1} r^{\mu_t}_t\, \bigg|\, \pi \bigg].
\end{align}
where the expectation is taken over trajectories of full episodes $\tau$, and $|\tau|$ is the length of episode $\tau$.

The objective is then to find an architecture for $\pi$ that is able to express strategies that perform the best according to Eq.~\ref{eq:test.performance} and, at the same time, can be efficiently learned even for moderately short training phases.


\begin{figure}[t]
\vskip 0.35in
	\begin{minipage}[t]{0.47\columnwidth}
		\centering
		\begin{tikzpicture}[scale=0.5]
		\draw[step=1cm,gray,very thin] (0,0) grid (5,6);
		\node[font=\scriptsize] at (0.5, 5.5) {G1};
		\node[font=\scriptsize] at (4.5, 5.5) {G2};
		\draw[fill=Cyan, opacity=0.2] (2,1) rectangle (3,2);
		\draw[fill=Orange, opacity=0.2] (2,0) rectangle (3,1);
		\draw[fill=ForestGreen, opacity=0.2] (0,5) rectangle (1,6);
		\draw[fill=ForestGreen, opacity=0.2] (4,5) rectangle (5,6);
		\node[font=\tiny] at (2.5, 0.5) {sign};
		\node[font=\tiny] at (2.5, 1.5) {start};
		\end{tikzpicture}
	\end{minipage}
	\hfill
	\begin{minipage}[t]{0.52\columnwidth}
		\vspace{-1.35in}
		\caption{An environment with two tasks: the goal location ($G1$ or $G2$) changes at each episode. The sign reveals the location of the goal. Optimal informed policies are shortest paths from \emph{start} to either $G1$ or $G2$, which never visit the sign. Thompson sampling cannot represent the optimal exploration/exploitation policy (go to the sign first) since going to the sign is not feasible by any informed policy.
		}
		\label{fig:maze1}
	\end{minipage}
\end{figure}
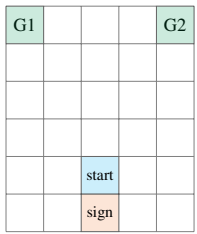

\section{Related Work and Contributions}
\label{sec:related_work}

In this section, we review how the online adaptation setting has been tackled in the literature. The main approaches are depicted in Fig.~\ref{fig:architecture}. We first compare the different methods in terms of the optimality of strategies they can represent, how they deal with non-stationary environments, and whether they leverage the efficient learning of informed policies. We then discuss learning task embeddings and how the various methods deal with unknown or irrelevant task descriptors. The last subsection summarizes our contributions. Unless otherwise stated, the methods described below have not been tested in the non-stationary setting. 

\subsection{Online Adaptation with Deep RL}
In the previous section we mentioned that the best exploration strategy corresponds to the optimal policy of the associated POMDP. Since the belief state $b_t(\tau_t)$ is a sufficient statistic of the history, POMDP policies takes the form $\pi(a_t|\tau_t) = \pi(a_t| s_t, b_t)$. 
While it is impractical to compute the exact belief state even for toy discrete problems, approximations can be learnt using Recurrent Neural Networks (RNNs) \citep{Bakker2001LSTM, control_rnn, Duan2016RL2}.
RNN-policies are directly trained to maximize the cumulative reward and do not leverage information about task descriptors at train time. While this class of policies can represent rich exploratory strategies, their large training complexity makes them highly suboptimal whenever the training phase is too short.

In order to reduce the training complexity of RNN policies, existing strategies have constrained the set of possible exploratory behaviors by leveraging privileged information about the task. 
Probe-Then-Exploit (PTE)~\citep[\eg][]{lattimore2018bandit,zhou2019probing} works in two phases.
First, it executes a pure exploratory policy with the objective of identifying the underlying task (\ie $\mu$) as accurately as possible. In the second phase, PTE runs the optimal policy associated to the estimated task. In that case, the agent only needs to learn a probing policy that maximizes the likelihood of the task at the end of the exploration phase and the informed policies for the training tasks, thus leading to a much more efficient training process. 
PTE has two main limitations. First, similarly to explore-then-commit approaches in bandits~\citep{Garivier2016explorethencommit}, the exploration can be suboptimal because it is not reward-driven: valuable time is wasted to estimate unnecessary information. Second, the switch between probing and exploiting is difficult to tune and problem-dependent, which makes these approaches unsuitable in non-stationary environments.

Thompson Sampling (TS)~\citep{thompson1933likelihood} leverages randomization to efficiently mix exploration and exploitation. 
Similarly to the belief state of an RNN-policy, TS maintains a distribution over tasks that are compatible with the observed history. 
The policy samples a task from the posterior and executes the corresponding informed policy for several steps. 
In that case, the training is limited to learning a maximum likelihood estimator to map trajectories to distributions over states and, similar to PTE, informed policies for the training tasks. This strategy proved successful in a variety of problems~\citep[\eg][]{chapelle2011empirical,Osband2017why, Osband2019deepexpl}. However, TS cannot represent certain probing policies because it is constrained to executing informed policies. 
Another drawback of TS approaches is that even in stationary environemnts, the frequency of re-sampling needs to be carefully tuned. This makes the application of TS to non-stationary environments challenging.
We describe an example of environment where TS is suboptimal in Fig.~\ref{fig:maze1}.

The Task Inference (\bnp) approach \citep{task_inference} 
is an RNN trained to simultaneously learn a good policy and predict the task descriptor $\mu$. 
Denoting by $m : H \to Z$ the mapping from histories to a latent representation of the belief state ($Z \subseteq \mathbb{R}^d$), the policy $\pi(a_t| z_t)$ selects the action based on the representation $z_t = m(\tau_t)$ constructed by the RNN. 
During training, $z_t$ is also used to predict the task descriptor $\mu_t$, using the \textit{task-identification} module $g: Z \to \mathcal{M}$.
The overall objective is:
\begin{align}
\mathbb{E} 
        \Big[\sum_{t=1}^{|\tau|}\gamma^{t-1} r^{\mu_t}_t\Big| \pi \Big] 
        + \underbrace{\beta \mathbb{E} 
        \Big[\sum_{t=1}^{|\tau|}\ell(\mu_t, g(z_t))\Big| \pi \Big] }_{\text{auxiliary loss}}
\end{align}
where $\ell(\mu_t, g(z_t))$ is the log-likelihood of $\mu_t$ under distribution $g(z_t)$. Note that because the auxiliary loss is only supposed to structure the memory of the RNN $m$ rather than be an additional reward for the policy, the gradient of the auxiliary loss with respect to $m$ ignores the effect of $m$ on $\pi$: given trajectories sampled by $\pi$, only the average gradient of $\ell$ with respect to $m$ is backpropagated.

Thus, the training of $\pi$ in \bnp is purely reward-driven and it does not suffer from the suboptimality of PTE/TS. However, in contrast to PTE/TS, it does not leverage the smaller sample complexity of training informed policies, and the auxiliary loss is defined over the whole value of $\mu$ while only some dimensions may be relevant to solve the task.

In the non-stationary setting, only a few models have been proposed, mainly based on the MAML algorithm. For instance, \cite{maml_adapt_online} combines MAML with model-based RL by meta-learning a transition model that helps an MPC controller predicting the action sequence to take. The method does not make use of the value of $\mu$ at train time and is specific to MPC controllers.

\subsection{Learning Task Embeddings}

The approaches described above differ in the necessary prior assumptions on tasks and their descriptions used for learning. The minimal requirement for all these approaches is to have access to \emph{task identifiers}, \ie one-hot encodings of the task. In general however, these approaches are sensitive to the description of the task. In particular, irrelevant variables have a significant impact on PTE approaches since the probing policy aims at identifying the task: for instance, an agent might waste time reconstructing the full map of a maze when it only needs to find a specific information to act optimally w.r.t the reward. Moreover, many methods are guided by a prior distribution over $\mu$ that has to be chosen by hand to fit the ground-truth distribution of tasks.

Several approaches have been proposed to learn from task identifiers~\citep{Gupta2018structured, Rakelly2019offpolicy,Zintgraf2019vate,Hausman2018LatentSpace}. The usual approach is to train embeddings of task identifiers jointly with the informed policies. TI is not amenable to joint task embedding training, since tasks are used as targets and not as inputs. \citet{task_inference} mention using TI with task embeddings, but the embeddings are pre-trained separately, which requires either additional interactions with the environment or expert traces.

\subsection{Contributions}
\label{subsec:contributions}

As for RNN/TI, IMPORT learns an RNN policy to maximize cumulative reward. As such, our approach does not suffer from the intrinsic limitations of PTE/TS in terms of optimality; because there is no decoupling between exploration and exploitation (of an informed policy), the approach does not suffer from the scheduling difficulties of PTE/TS and is  readily applicable to non-stationary environments. Nonetheless, similarly to PTE/TS and contrarilty to TI, IMPORT leverages fast training of informed policies through a joint training of an RNN and an informed policy. 

In addition, IMPORT does not rely on probabilistic models of task descriptors. Learning task embeddings makes the approach robust to irrelevant task descriptors contrary to \bnp, but also makes IMPORT applicable when only task identifiers are available.

The next section describes these components in more details.

\section{Method}
\label{sec:method}

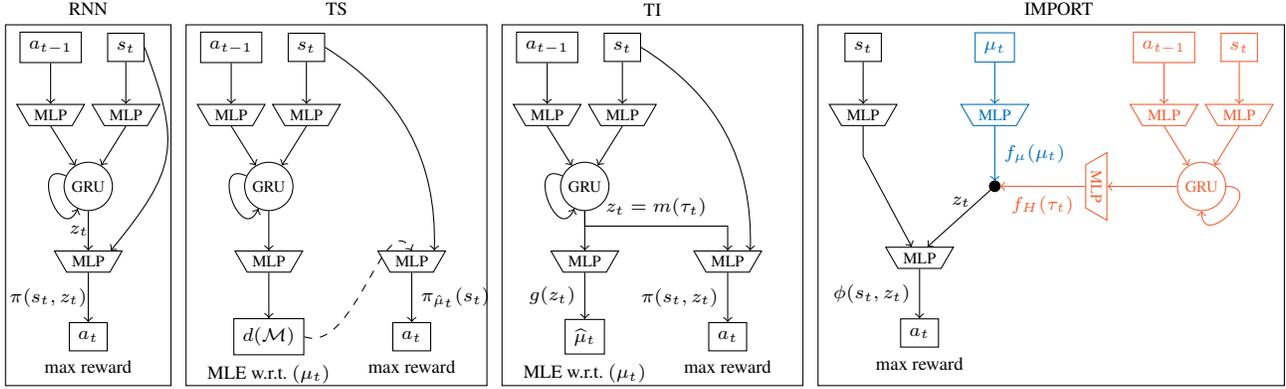
\begin{figure*}
        \centering
        \begin{tikzpicture}[font=\scriptsize]

                \begin{scope}
                \node[rectangle, draw] (atm1) at (0,5) {$a_{t-1}$};
                \node[rectangle, draw] (st) at (1,5) {$s_t$};

                \node[trapezium, trapezium left angle=120, trapezium right angle=120,
                        draw, font=\tiny, inner sep=2pt, below of=atm1, node distance=.9cm] (atmlp) {MLP};
                \node[trapezium, trapezium left angle=120, trapezium right angle=120,
                        draw, font=\tiny, inner sep=2pt, below of=st, node distance=.9cm] (stmlp) {MLP};
                \draw[->] (st.south) -- (stmlp.north);
                \draw[->] (atm1.south) -- (atmlp.north);
                \node[circle, draw, font=\tiny, inner sep=2pt, below of=atmlp, node distance=.95cm, xshift=.5cm] (lstm) {GRU};
                \draw[->] (stmlp.south) -- (lstm);
                \draw[->] (atmlp.south) -- (lstm);
                \path[->] (lstm) edge [loop above, out=155, in=245,looseness=3] (lstm);
                \node[trapezium, trapezium left angle=120, trapezium right angle=120,
                        draw, font=\tiny, inner sep=2pt, below of=lstm, node distance=1cm] (fhmlp) {MLP};
                \draw[->] (lstm.south) -- (fhmlp.north) node[midway, xshift=-3.5pt] {$z_t$};
                \node[rectangle, draw, below of=fhmlp, node distance=1cm] (act) {$a_t$};
                \draw[->] (fhmlp.south) -- (act.north) node[midway, left, xshift=3pt] {$\pi(s_t,z_t)$};
                \path[->] (st.east) edge[bend left, looseness=1.3] ($(fhmlp.north)+(0.3,0)$);
                \node[anchor=north] at (act.south) {max reward};
                \draw (-.6, 5.3) rectangle (1.6,0.5);
                \node[anchor=south] at (0.5, 5.3) {RNN};
                \end{scope}


                           \begin{scope}[shift={(.14\textwidth,0)}]
                \node[rectangle, draw] (atm1) at (0,5) {$a_{t-1}$};
                \node[rectangle, draw] (st) at (1,5) {$s_t$};

                \node[trapezium, trapezium left angle=120, trapezium right angle=120,
                        draw, font=\tiny, inner sep=2pt, below of=atm1, node distance=.9cm] (atmlp) {MLP};
                \node[trapezium, trapezium left angle=120, trapezium right angle=120,
                        draw, font=\tiny, inner sep=2pt, below of=st, node distance=.9cm] (stmlp) {MLP};
                \draw[->] (st.south) -- (stmlp.north);
                \draw[->] (atm1.south) -- (atmlp.north);
                \node[circle, draw, font=\tiny, inner sep=2pt, below of=atmlp, node distance=.95cm, xshift=.5cm] (lstm) {GRU};
                \draw[->] (stmlp.south) -- (lstm);
                \draw[->] (atmlp.south) -- (lstm);
                \path[->] (lstm) edge [loop above, out=155, in=245,looseness=3] (lstm);
                

                \node[trapezium, trapezium left angle=120, trapezium right angle=120,
                        draw, font=\tiny, inner sep=2pt, below of=lstm, node distance=1cm] (mumlp) {MLP};
                \node[rectangle, draw, below of=mumlp, node distance=1cm] (mudistrib) {$d(\mathcal{M})$};
                \draw[->] (lstm.south) -- (mumlp.north); 
                \draw[->] (mumlp.south) -- (mudistrib.north) node[midway, left] {};
                \node[anchor=north] at (mudistrib.south) {MLE \wrt $(\mu_t)$};

                \node[trapezium, trapezium left angle=120, trapezium right angle=120,
                        draw, font=\tiny, inner sep=2pt, below of=lstm, node distance=1cm, xshift=1.9cm] (fhmlp) {MLP};
                \coordinate (yx) at ($(lstm.south)+(0, -0.2)$);
          
                \node[rectangle, draw, below of=fhmlp, node distance=1cm] (act) {$a_t$};
                \path[->] (st.east) edge[bend left] ($(fhmlp.north)+(0.3,0)$);
                 \path[dashed, ->] (mudistrib.east)  edge[out=-20, in=120]  (fhmlp.north);
                \draw[->] (fhmlp.south) -- (act.north) node[midway, right] {$\pi_{\hat{\mu}_t}(s_t)$};
                \node[anchor=north] at (act.south) {max reward};
                
                \draw (-.6, 5.3) rectangle (3.4,0.5);
                \node[anchor=south] at (1.4, 5.3) {TS};
                \end{scope}

                \begin{scope}[shift={(.385\textwidth,0)}]
                \node[rectangle, draw] (atm1) at (0,5) {$a_{t-1}$};
                \node[rectangle, draw] (st) at (1,5) {$s_t$};

                \node[trapezium, trapezium left angle=120, trapezium right angle=120,
                        draw, font=\tiny, inner sep=2pt, below of=atm1, node distance=.9cm] (atmlp) {MLP};
                \node[trapezium, trapezium left angle=120, trapezium right angle=120,
                        draw, font=\tiny, inner sep=2pt, below of=st, node distance=.9cm] (stmlp) {MLP};
                \draw[->] (st.south) -- (stmlp.north);
                \draw[->] (atm1.south) -- (atmlp.north);
                \node[circle, draw, font=\tiny, inner sep=2pt, below of=atmlp, node distance=.95cm, xshift=.5cm] (lstm) {GRU};
                \draw[->] (stmlp.south) -- (lstm);
                \draw[->] (atmlp.south) -- (lstm);
                \path[->] (lstm) edge [loop above, out=155, in=245,looseness=3] (lstm);
                
                \node[trapezium, trapezium left angle=120, trapezium right angle=120,
                        draw, font=\tiny, inner sep=2pt, below of=lstm, node distance=1cm] (mumlp) {MLP};
                \node[rectangle, draw, below of=mumlp, node distance=1cm] (muhat) {$\widehat{\mu}_t$};
                \draw[->] (lstm.south) -- (mumlp.north); 
                \draw[->] (mumlp.south) -- (muhat.north) node[midway, left] {$g(z_t)$};
                \node[anchor=north] at (muhat.south) {MLE \wrt $(\mu_t)$};

                \node[trapezium, trapezium left angle=120, trapezium right angle=120,
                        draw, font=\tiny, inner sep=2pt, below of=lstm, node distance=1cm, xshift=1.9cm] (fhmlp) {MLP};
                \coordinate (yx) at ($(lstm.south)+(0, -0.2)$);
                \draw[->] (lstm.south) -- (yx) -- (yx -| fhmlp.north) node[midway, above] {$z_t = m(\tau_t)$} -- (fhmlp.north);
                \node[rectangle, draw, below of=fhmlp, node distance=1cm] (act) {$a_t$};
                \path[->] (st.east) edge[bend left] ($(fhmlp.north)+(0.3,0)$);
                \draw[->] (fhmlp.south) -- (act.north) node[midway, left] {$\pi(s_t, z_t)$};
                \node[anchor=north] at (act.south) {max reward};
                
                \draw (-.6, 5.3) rectangle (3.4,0.5);
                \node[anchor=south] at (1.4, 5.3) {\bnp};
                \end{scope}

                \begin{scope}[shift={(.63\textwidth, 0)}]
                \node[rectangle, draw] (st) at (0,5) {$s_t$};
                \node[rectangle, draw, NavyBlue] (mut) at (1.75,5) {$\mu_t$};
                \node[rectangle, draw, RedOrange] (at) at (4,5) {$a_{t-1}$};
                \node[rectangle, draw, RedOrange] (st2) at (5,5) {$s_t$};
                \node[trapezium, trapezium left angle=120, trapezium right angle=120,
                        draw, font=\tiny, inner sep=2pt, below of=st, node distance=.9cm] (stmlp) {MLP};
                \node[trapezium, trapezium left angle=120, trapezium right angle=120, NavyBlue,
                        draw, font=\tiny, inner sep=2pt, below of=mut, node distance=.9cm] (mutmlp) {MLP};
                \draw[->] (st.south) -- (stmlp.north);
                \draw[->, NavyBlue] (mut.south) -- (mutmlp.north);
                \node[trapezium, trapezium left angle=120, trapezium right angle=120,
                        draw, font=\tiny, inner sep=2pt, below of=stmlp, node distance=1.9cm, xshift=.75cm] (phimlp) {MLP};
                
                \coordinate (fmu) at ($(mutmlp.south)+(0, -0.8cm)$); 
                \draw[->] (stmlp.south) -- +(0, -0.4cm) -- ($(phimlp.north)-(3pt,0)$);
                \node[rectangle, draw, below of=phimlp, node distance=1cm] (act) {$a_t$};
                \draw[->] (phimlp.south) -- (act.north) node[midway, left] {$\phi(s_t, z_t)$};
                \draw[fill=black] (fmu) circle(2pt);
                \draw[->, NavyBlue] (mutmlp.south) -- ($(fmu)+(0,2pt)$) node[midway, right, NavyBlue] {$f_\mu(\mu_t)$};
                \draw[->] (fmu) -- ($(phimlp.north)+(3pt,0)$) node[midway, above] {$z_t$};

                \node[trapezium, trapezium left angle=120, trapezium right angle=120, RedOrange,
                        draw, font=\tiny, inner sep=2pt, below of=at, node distance=.9cm] (atmlp) {MLP};
                \node[trapezium, trapezium left angle=120, trapezium right angle=120, RedOrange,
                        draw, font=\tiny, inner sep=2pt, below of=st2, node distance=.9cm] (st2mlp) {MLP};
                \draw[->, RedOrange] (st2.south) -- (st2mlp.north);
                \draw[->, RedOrange] (at.south) -- (atmlp.north);
                \node[circle, RedOrange,
                        draw, font=\tiny, inner sep=2pt, below of=atmlp, node distance=.95cm, xshift=.5cm] (lstm) {GRU};
                \draw[->, RedOrange] (st2mlp.south) -- (lstm);
                \draw[->, RedOrange] (atmlp.south) -- (lstm);
                \node[trapezium, trapezium left angle=120, trapezium right angle=120, RedOrange,
                        draw, font=\tiny, inner sep=2pt, left of=lstm, node distance=1.4cm, rotate=-90] (fhmlp) {MLP};
                \draw[->, RedOrange] (lstm.west) -- (fhmlp.north);
                \draw[->, RedOrange] (fhmlp.south) -- ($(fmu)+(2pt,0)$) node[midway, below] {$f_H(\tau_t)$};
                \path[->, RedOrange] (lstm) edge [loop above, out=0, in=-90,looseness=3] (lstm);
                \node[anchor=north] at (act.south) {max reward};
                \draw (-.6, 5.3) rectangle (5.6,0.5);
                \node[anchor=south] at (2.6, 5.3) {IMPORT};

                \end{scope}
        \end{tikzpicture}
        \caption{Representation of the different architectures. RNN policy is a classical recurrent neural network. Thompson Sampling (TS) samples a value of $\mu$ at each timestep and follows the corresponding informed policy. \bnp is a recurrent policy such that one can predict $\mu$ from the hidden state. At last, IMPORT is composed of two models sharing parameters: The (black+blue) architecture is the informed policy $\pi_\mu$ optimized through Eq. 1 \textbf{(B)} while the (black+red) architecture is the history-based policy $\pi_H$ (used at test time) trained through Eq 1. \textbf{(A)} (and eventually \textbf{(C)}).}
        \label{fig:architecture}
\end{figure*}

In this section, we describe the main components of the IMPORT model, as well as the online optimization procedure and an additional auxiliary loss to further speed-up learning. The overall approach is described in Fig.~\ref{fig:architecture} (right).

\subsection{Regularization Through Informed Policies}
Our approach leverages the knowledge of the task descriptor $\mu$ and informed policies to construct a latent representation of the task that is \emph{purely reward driven}.
Since $\mu$ is unknown at testing time, we use this informed representation to train a predictor based on a recurrent neural network. 
To leverage the efficiency of informed policies even in this phase, we propose an architecture \emph{sharing parameters} between the informed policy and the final policy such that the final policy will benefit from parameters learned with privileged information.
The idea is to constrain the final policy to stay close to the informed policy while allowing it to perform probing behaviors when needed to effectively reduce the uncertainty about the task. 
We call this approach InforMed POlicy RegularizaTion (IMPORT).

Formally, we define by $\pi_{\mu}(a_t|s_t, \mu)$ and $\pi_H(a_t| \tau_t)$ the informed policy and the history-dependent policy that will be used at test time.
The informed policy $\pi_\mu =  \phi  \circ f_\mu$ is defined as the functional composition of $f_\mu$ and $\phi$, where $f_\mu : \mathcal{M} \to Z$ projects $\mu$ in a latent space $Z \subseteq \mathbb{R}^k$ and $\phi: \mathcal{S} \times Z \to \mathcal{A}$ selects the action based on the provided latent representation. 
The idea is that $f_\mu(\mu)$ captures the relevant information contained in $\mu$ while ignoring dimensions that are not relevant for learning the optimal policy. This behavior is obtained by training $\pi_\mu$ directly to maximize the task reward $r_\mu$ (\ie informed).

While this policy leverages the knowledge of $\mu$ at training time, $\pi_H$ should be able to act based on the sole history.
To encourage $\pi_H$ to behave like the informed policy while preserving the ability to probe, 
we let $\pi_H$ share parameters with $\pi_\mu$ through the $\phi$ component that they have in common.
We thus define $\pi_H = \phi \circ f_H$ where $f_H : \mathcal{H} \to Z$ encodes the history into the latent space.
By sharing the policy head $\phi$ between informed and history-dependent policies, the approximate belief state constructed by the RNN is mapped to the same latent space as $\mu$. As such, when informed policies learn faster than the $\pi_H$, they can be reused directly by $\pi_H$ when the uncertainty about the task is small.

More precisely, let $\theta, \omega, \sigma$ the parameters of $\phi, f_H$ and $f_\mu$ respectively, so that $\pi_{\mu}^{\sigma \theta}(a_t|s_t, \mu_t) = \phi^{\theta} \circ f_\mu^{\sigma}  = \phi^\theta(a_t | s_t, f_{\mu}^{\sigma}(\mu_t))$ and $\pi_{H}^{\omega \theta}(a_t|\tau_t) =  \phi^{\theta} \circ f_H^{\omega}  = \phi^\theta(a_t | s_t, f_{H}^{\omega}(\tau_t))$. The goal of IMPORT is to maximize over  $\theta, \omega, \sigma$ the following objective function:
\begin{equation}
        \label{eq:import.fullopt}
        \begin{aligned}
        &\underbrace{\mathbb{E} 
        \Big[\sum_{t=1}^{|\tau|} \gamma^{t-1} r_{t}^{\mu_t}\, \Big|\, \pi_H^{\omega, \theta} \Big]}_{\textbf{(A)}}
        +\underbrace{ \lambda  \mathbb{E} 
        \Big[\sum_{t=1}^{|\tau|} \gamma^{t-1} r_{t}^{\mu_t}\, \Big|\,\pi_\mu^{\sigma, \theta}  \Big]}_{\textbf{(B)}}
        \end{aligned}
\end{equation}

\subsection{IMPORT with $\beta > 0 $: Speeding Up the Learning.}
The only information that is shared between  $\pi_H$ and $\pi_{\mu}$ is function $\phi$.
However, optimizing term \textbf{(B)} in Eq.~\ref{eq:import.fullopt}  produces also a reward-driven latent representation of the task through function $f_{\mu}$.
This information can be used to regularize the prediction of $f_H$, and to encourage the history-based policy to predict a task embedding close to the one predicted by the informed policy. We can thus rewrite Eq.~\ref{eq:import.fullopt} as:
\begin{align}
        & \mathbb{E}\Bigg[\sum_{t=1}^{|\tau|} \gamma^{t-1} r^{\mu_t}_{t} \,\bigg|\, \pi_H^{\omega, \theta} \bigg] + \lambda \mathbb{E}\bigg[\sum_{t=1}^{|\tau|} \gamma^{t-1} r^{\mu_t}_{t}\, \bigg|\, \pi_\mu^{\sigma, \theta} \bigg]\nonumber\\
        &
        \quad \underbrace{-\beta\mathbb{E}\Bigg[ \sum_{t=1}^{|\tau|} D\Big(f_\mu(\mu_t), f_H(\tau_t) \Big)\, \bigg|\, \pi_H^{\omega, \theta}\bigg]}_{\text{auxiliary loss}~~ \textbf{(C)}}        \label{eq:import.fullopt.reg}
\end{align}
where $D$ is the squared $2$-norm in our experiments. Note that the objective \textbf{(C)} is an auxiliary loss, so only the \textit{average} gradient of $D$ with respect to $f_H$ along trajectories collected by $\pi_H$ is backpropagated, ignoring the effect of $f_H$ on $\pi_H$.

\begin{algorithm}[t]
        \caption{IMPORT Training}
        \label{alg:alternating}
        \footnotesize
        \begin{algorithmic}
                \STATE Initialize $\sigma, \omega, \theta$ arbitrarily
                \FOR{$n = 1, \ldots, N$}
                \STATE Collect $M$ trajectories according to $\pi_H$ in buffer $B_H$
                \STATE Collect $M$ trajectories according to $\pi_{\mu}$ in buffer $B_{\mu}$
                \STATE Compute $\nabla_{\omega,\theta}$\textbf{(A)} on $B_H$
                \STATE Compute average $\nabla_{\omega} D$ on $B_H$ (auxiliary loss \textbf{(C)})
                \STATE Compute $\nabla_{\sigma,\theta}$\textbf{(B)} on $B_\mu$
                \STATE Update $\sigma,\theta$ and $\omega$ based on $\nabla_{\sigma,\theta}$\textbf{(A)}$+\nabla_{\omega,\theta}$\textbf{(B)} + $\nabla_{\omega} D$
                \ENDFOR
        \end{algorithmic}
\end{algorithm}

\subsection{Optimization}
We propose an optimization scheme for \eqref{eq:import.fullopt} and \eqref{eq:import.fullopt.reg} based on policy gradient algorithms (our experiments use A2C \citep{Mnih2016A3C}  and REINFORCE \cite{Baxter2001gpomdp}). The high-level algorithm is summarized in Alg. \ref{alg:alternating}. At each iteration, we collect two batches of episodes, one containing $M$ trajectories (full episodes) generated by the exploration/exploitation policy $\pi_H$, and the other one containing $M$ trajectories generated by the informed policy  $\pi_\mu$. The gradients of the RNN ($f_H$) and of the policy head $(\phi)$ are computed on the first batch according to objectives \textbf{(A)} and \textbf{(C)} (if $\beta>0$ in Eq. \eqref{eq:import.fullopt.reg}), while the gradients of the policy head ($\phi$) and of the task embeddings ($f_{\mu}$) are computed on the second batch. The weight updates are performed on the sum of all gradients. 

\section{Experiments}
\label{sec:experiments}

\begin{table*}[t]
 \centering
    \begin{tabular}{lrrrrrr}\toprule
     \textbf{Method} &  \multicolumn{6}{c}{ \textbf{Test reward} }\\ \cmidrule(r){2-7}
    & \multicolumn{3}{c}{ $\mu_{max}$ = 0.9 ($\mu_{min}=0.1$)} & \multicolumn{3}{c}{ $\mu_{max}$ = 0.5 ($\mu_{min}=0.1$)} \\\cmidrule(r){2-7}
    &  $K=5$ & $K=10$ & $K=20$  &  $K=5$ & $K=10$ & $K=20$  \\\midrule
    UCB & 64.67(0.38) & 52.03(1.13) &  37.61(1.5) & 31.12(0.21) & 22.84(0.97) & \textbf{15.63(0.74)} \\
    SW-UCB & 62.17(0.74) & 50.8(1.21) & 33.02(0.19) & 30.06(1.36) & 22.11(0.69) & 14.63(0.57)\\
     TS  &68.93(0.7) &  41.35(0.98) & 20.58(1.86) & 28.93(1.23) & 15.23(0.28) & 9.8(0.64)\\
    RNN & \textbf{73.39(0.78)} &  54.21(2.94) &  30.51(0.8) & 31.63(0.12) & 21.19(0.87) & 11.81(0.94) \\
    TI& 73.01(0.86) &  58.81(1.67) & 32.22(0.89) & 31.9(0.82) & 21.46(0.37) & 12.78(0.77) \\
    IMPORT($\beta=0$)& 72.66(0.96) & 57.62(0.75) & 31.36(4.55)& 30.76(0.7) & 21.53(2.07)  & 11.41(0.46) \\
    IMPORT($\beta>0$) & 73.13(0.62) & \textbf{61.38(0.62)} & \textbf{42.29(3.08)} & \textbf{32.44(1.12} & \textbf{24.83(0.87)}& 14.47(2.05) \\
        \bottomrule
    \end{tabular}
    \caption{Bandits in non stationary modes where $\rho=0.05$ and $T=100$. On average, the distribution over the arms changes 5 times per episode. The number of arms is $K$. Each value corresponds to the cumulated reward after $4 \cdot 10^7$ environment steps (and standard deviation over the 3 seeds). }
    \label{tab:bandit}
\end{table*}

\subsection{Experimental Setting and Baselines}

\begin{table}[t]
\small{
 \centering
    \begin{tabular}{lrrr}\toprule
     \textbf{Method} &  \multicolumn{3}{c}{ \textbf{Test reward} }\\ \cmidrule(r){2-4}
     & $\rho = 0$ & $\rho = 0.05$  & $\rho=0.1$\\\midrule
    TS & 93.54(1.99) & 85.33(6.28) & 91.34(4.28) \\
    RNN & 81.3(12.7) & 73.6(6.3) & 76(4.1) \\
    TI & 90.4(0.5) & 83.6(2.4) & 80.1(2.2) \\
    IMPORT($\beta=0$)& 90.66(6.6) & 69.1(9.1) & 79(2.5) \\
    IMPORT($\beta>0$) & \textbf{96.4(2.9)} & \textbf{93.4(3.2)} & \textbf{97.2(1.5)} \\
        \bottomrule
    \end{tabular}
    }
    \caption{CartPole stationary and non-stationary where $T=100$ (after $10^7$ environment steps).}
    \label{tab:cp}
\end{table}

\begin{table}[t]
 \begin{center}
\small{
    \begin{tabular}{lrr}\toprule
     \textbf{Method} &  \multicolumn{2}{c}{ \textbf{Test reward} }\\ \cmidrule(r){2-3}
     & $\rho = 0.01$  & $\rho=0.05$\\\midrule
    RNN &  -406.4(30.1) & -459.(57.5)  \\
    TI &  -278.4(45.)  & -223.6(36.3) \\
    TS  &   -242.9(4.9) & -190.4(51.9) \\
    IMPORT &  \textbf{-112.2(10.8)} & \textbf{-101.7(11.1)}  \\
        \bottomrule
    \end{tabular}
    }
    \end{center}
    \caption{Acrobot in non-stationary settings with $T=500$ (after $6 \cdot 10^6$ environment steps).}
    \label{tab:acro}
\end{table}

\begin{table}[t]
\small{
 \centering
    \begin{tabular}{lrrrr}\toprule
     \textbf{Method} &  \multicolumn{4}{c}{ \textbf{Test reward} }\\ \cmidrule(r){2-5}
    & \multicolumn{2}{c}{ $R=10$} & \multicolumn{2}{c}{$R=100$} \\\cmidrule(r){2-5}
    &  $\rho = 0$ &  $\rho = 0.1$  &  $\rho = 0$ &  $\rho = 0.1$  \\\midrule
    TS  & 72.2(15.4) & 44.9(8.2)& 89.8(2.2) & 49.2(4.)\\
    RNN & 81.4(9.7) & 74.3(8.4) &79.8(9.4) &63.2(6.)\\
    TI &  \textbf{88.6(4.4)} &   80.8(5.) & 85.2(7.5)& 71.2(8.4)\\
    IMPORT & 77.1(17.8) & \textbf{89.4(6.)} & \textbf{ 90.1(1.5)} &  \textbf{  88.3(1.)} \\
        \bottomrule
    \end{tabular}
    }
    \caption{CartPole: Generalization with Task Embeddings. $R$ is the number of training tasks, while the models have been selected on 100 different validation tasks, and the performance is reported on 100 test tasks. $\mu$ is a one-hot vector of size $R$ encoding the id of each training task (after $10^7$ environment steps).}
    \label{tab:cp2}
\end{table} 

Our model is compared to different baselines:
\textbf{Recurrent Neural Networks (RNN)} is a recurrent policy based on a GRU recurrent module \cite{control_rnn} that does not use $\mu$ at train time but just the observations and actions. \textbf{Thompson Sampling (TS)} and \textbf{Task Inference (\bnp)} models are trained in the same setting that the \textbf{IMPORT} model, using $\mu$ at train time, but not at test-time. 

\begin{figure}[t]
\centering
\includegraphics[width=1.0\linewidth]{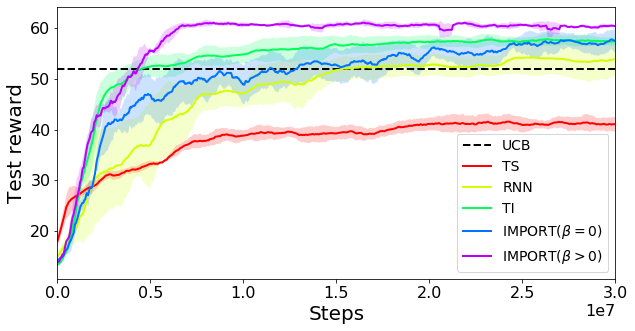}
\caption{Training curves on the bandit problem where $K=10$, $p_{max}=0.9$ and $\rho=0.05$}
\label{fig:bandits}
\end{figure}

\begin{figure}[t]
\centering
\includegraphics[width=1.0\linewidth]{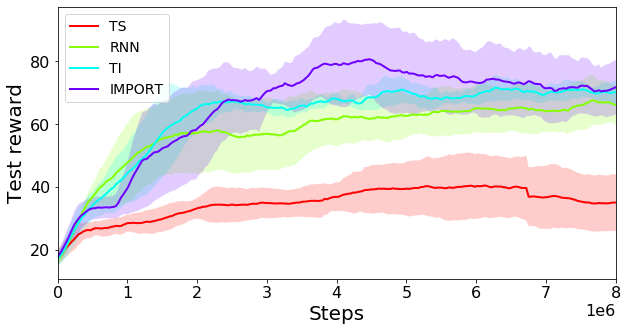}
\caption{Cartpole with $R=10$ training tasks and  $\rho=0.1$. The curves have been selected on 100 validation tasks, and the results are reported over 100 test tasks unseen at train time.}
\label{fig:cartpole_task}
\end{figure}

\begin{figure}[t]
\centering
\includegraphics[width=0.9\linewidth]{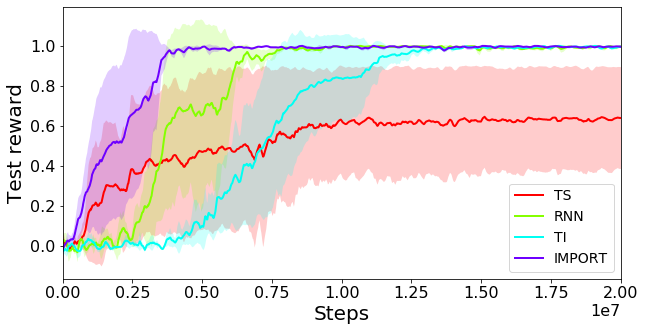}
\includegraphics[width=0.9\linewidth]{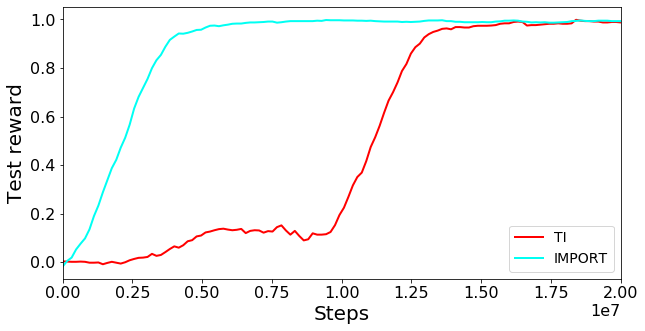}
\caption{Training curves on Maze2d (up) and Maze3d (down) environments under scenario A learned with REINFORCE. 
The curves show the ability of IMPORT to learn faster than TI and RNN. TS struggles on such problems because of its oscillating behaviour, however, under scenario A, it can touch the sign. Under scenario B, TS is deemed to fail.}
\label{fig:maze}
\end{figure}

Each approach (\ie TS, RNN, TI and IMPORT) is trained using A2C\footnote{REINFORCE has been used for Maze problems because it was more stable in this sparse reward setting.} \cite{Mnih2016A3C} on $10$ CPUs\footnote{+ 1 GPU in the Maze 3d setting}. 
Precise values of the hyper-parameters and details on the neural network architectures are also given in the appendix \ref{sub:network_archi} . The networks for all approaches have a similar structure with the same number of hidden layers and same hidden layers sizes for fair comparison.
In all the tables, we report the test performance of the best hyper-parameter value, selected according to the procedure described in appendix \ref{ap:results_preprocessing} 

\subsection{Environments}
 Experiments have been performed on a diverse set of environments whose goals are to showcase different aspects: a) generalization to unseen tasks, b) adaptation to non-stationarities, c) ability to learn complex probing policies, d) coping with high-dimensional state space and e) sensitivity to the task descriptor.

In both stationary and non-stationary settings, we study multi-armed bandits (MAB) and control tasks (CartPole and Acrobot) to test a) and b) characteristics. They all have high dimensional task descriptors that contain irrelevant variables, thus potentially sensitive to reconstructing unnecessary information e). \\
We then provide two environments with sparse rewards but different state inputs: Maze2d with $xy$ coordinates, and Maze3d is a challenging first-person view task with high dimensional state space (pixels) to  test d). Maze environments require c) to discover complex probing policies.\\
Note that in MAB, CartPole and Acrobot, train/validation/test task sets are disjoint to support a).
To test e), we study CartPole for two types of privileged information: 1) $\mu$ summarizes the dynamics of the system, e.g. pole length and 2) $\mu$ encodes a task index (one-hot vector) in a set of training tasks. When using 2), we restrict the size of the training set to be small, thus making generalization property a) even harder. 

Environments are described in further details in Appendix \ref{ap:experiments}.

\paragraph{Multi-armed bandits: } We use MAB problems with $K$ arms where each arm $k$ is Bernoulli with parameter $\mu_k$. In the non-stationary case, there is a probability $\rho$ to re-sample the value of $\mu=(\mu_k)_{k=1}^{K}$ at each timestep,  The size of each episode is $T=100$. We consider the setting where one single random arm is associated with a Bernoulli of parameter $\mu_{max}$, while the other arms are associated with Bernoulli with parameters independently drawn uniformly in $[0,\mu_{min}]$. In this environment, TS and TI are using a \textit{Beta} distribution to model $\mu$. We compare also with some bandit-specific online algorithms: \textit{UCB}~\citep{Auer2003UCB} and one of its non-stationary counter-part \textit{Sliding-Window UCB} ~\citep{window_ucb}. 

\paragraph{Control Environments: } We use two environments, \textit{CartPole} and \textit{Acrobot}, where $\mu$ is controlling different physical variables of the system, e.g., the weight of the cart, the size of the pole, etc. The size of $\mu$ is $5$ for Cartpole and $7$ for Acrobot. The values of $\mu$ components are normalized between $-1$ and $+1$, are uniformly sampled at the beginning of each episode, and can be resampled at each timestep with a probability $\rho$. The maximum size of each trajectory is $T=100$ for Cartpole and $T=500$ for Acrobot, and the reward functions are the one implemented in OpenAI Gym \cite{1606.01540}. These environments are particularly difficult for two reasons: the `direction' of the forces applied to the system may vary such that the optimal informed policies are very different w.r.t.\ to $\mu$. Moreover, the mapping between $\mu$ and the dynamics is not obvious since some dimensions of $\mu$ may be irrelevant, and some others may have opposite or similar effects. In these environments, TS and TI are using a \textit{Gaussian} distribution to model $\mu$.
CartPole with a task identifier as $\mu$ is described in \ref{subsec:results} and Appendix \ref{ap:cartpole_task}.

\paragraph{Maze2d/Maze3d with Two Goals and Sign: }

\begin{figure}[t]
\centering
\includegraphics[width=1.0\linewidth]{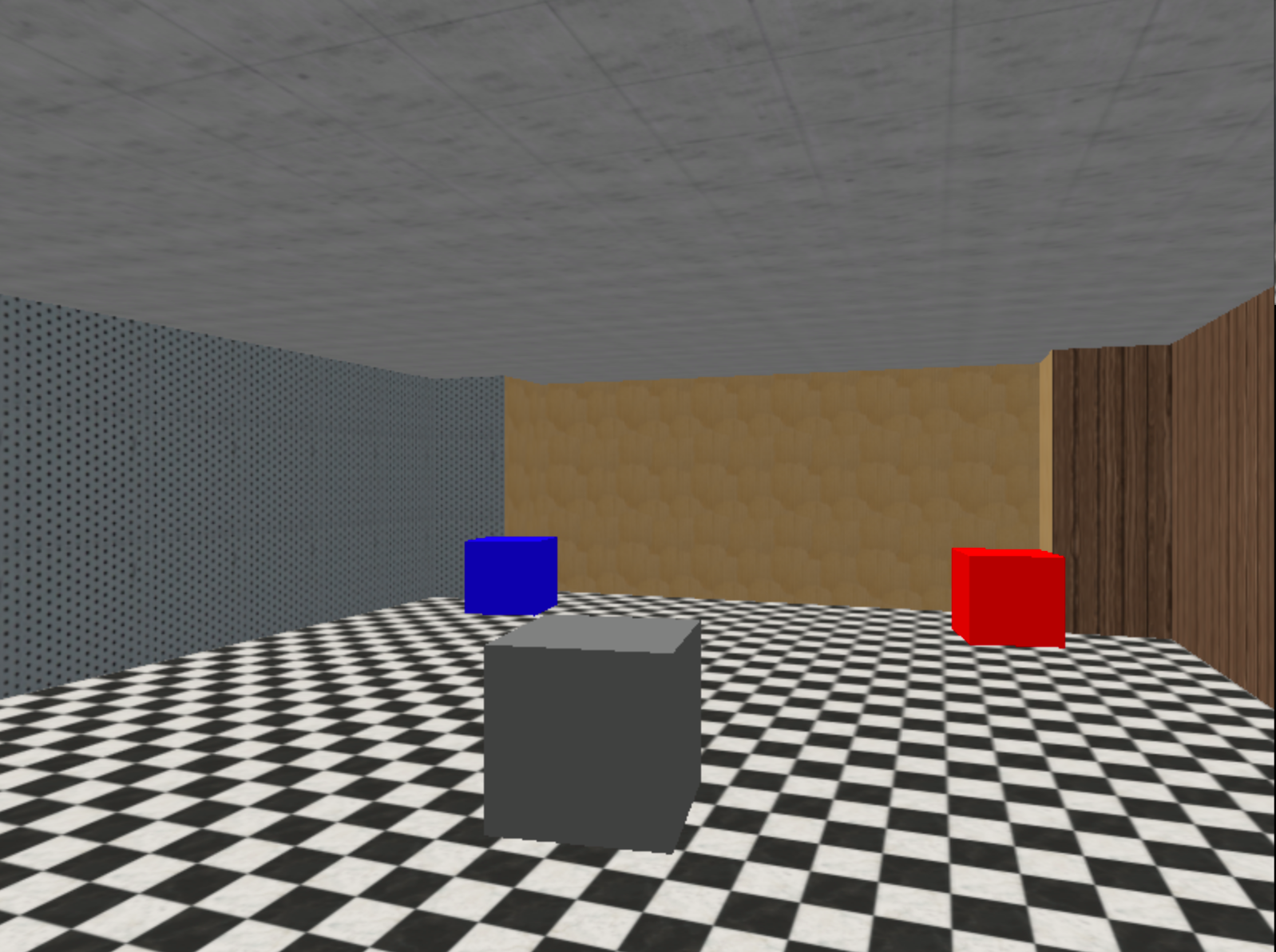}
\caption{Maze 3d under scenario A: The goal is either located at the blue or the red box. The grey box is the sign that provide the goal location when the agent is close. The optimal policy (i.e. reaching the sign then the right goal location) has an average length of 19 steps. }
\label{fig:maze_layout}
\end{figure}

We consider (see Fig. \ref{fig:maze_layout}) an environment where two possible goals are positioned at two different locations. The value of $\mu=+1$ or $-1$ denotes which of the two goals is active and it is sampled at random at the beginning of each episode. The agent can access this information by moving to a \textit{sign} location. It has to reach the goal in at most $100$ steps -- the optimal policy is able to achieve this objective in 19 steps -- to receive a reward of $+1$ and the episode stops. If the agent reaches the wrong goal, it receives a reward of $-1$ and the episode stops. The agent has three actions (forward, turn left, turn right) and the environment has been implemented using \emph{MiniWorld}~\citep{gym_miniworld}. We propose two versions: the 2d version where the agent observes its location (and eventually the goal location when going over the \emph{sign}), and the 3d version where the agent observes an image, the sign information is encoded as a fourth dimension (either zeros or $\mu$ values when the sign is touched) concatenated to the RGB image. Note that it is an environment with a sparse reward that is very complex to solve with a RNN which has to learn to memorize the \emph{sign} information. In this setting, TS and TI are using a \textit{Bernoulli} distribution to model $\mu$. 
We tried two scenarios A and B detailed in Appendix \ref{ap:maze} with difficulty adjusted with the sign position: in the scenario A, the sign is between the initial position and the two possible goals, whereas in scenario B it is behind the initial position as in Fig. \ref{fig:maze1}.

\subsection{Results}
\label{subsec:results}

Table  \ref{tab:bandit}, \ref{tab:cp}, \ref{tab:acro} and \ref{tab:cp2} show the cumulated reward obtained by the different methods over the different environments, with the standard deviation over the 3 seeds. 

\textit{Quality of the learned policy: } In all environments, the \bnp and IMPORT methods are performing better or similarly to the baselines (TS and RNN). Indeed \bnp and IMPORT are able to benefit from the values of $\mu$ at train time to avoid poor local minima and to learn a good policy in a reasonable time. It can be seen that the ordering of the methods depends on the environment: in control problems, TS is performing well since the dynamics of the environment can be captured in a few transitions and the informed policy acts as a sufficient exploration policy, but is performing bad on other settings. The non-stationary settings make the different problems more difficult to most of the methods, but IMPORT suffers less of this non-stationarity than other methods.

\textit{Learning Speed: } One important aspect to look at is the ability to learn with few interactions with the environment. Indeed, as explained before, RNN, \bnp and IMPORT have the same expressiveness -- i.e., their final policy is a recurrent neural network -- but are trained differently. Figure \ref{fig:bandits} \footnote{All the learning curves are provided in the appendix.} shows the best training curves (averaged over 3 seeds) for the different methods on one of the bandit setting. IMPORT is able to reach a better performance with less interactions with the environment than the baselines, and particularly than RNN which is not guided by the $\mu$ information at train time. Note that this is observed in almost all the experimental results on all the environments. When comparing IMPORT$(\beta=0)$ and IMPORT$(\beta>0)$, the learning speed of IMPORT$(\beta>0)$ is better since it benefits from an auxiliary loss of the same nature than TI. But coupled with the informed policy regularization principle, it achieves better and faster than TI. Note that the abilty of IMPORT to learn fast is particularly visible on the Maze experiments (see Figure \ref{fig:maze}). 
Depending on the scenario, TS may perform badly. This is mainly due to the fact that, $\phi$ is trained with $\mu$ as an input; it prevents the agent to actually follow probing actions, thus making the auxiliary supervised objective ineffective. For CartPole-task, TS predicts $\mu$ using a multinomial distribution; at test time, it behaves according to informed policies $\pi_{\hat{\mu}_t}$ where $ \hat{\mu}_t$ corresponds to the index of one training task.

\textit{Generalization and sensitivity to task descriptors: } We have performed a set of experiments to evaluate both generalization to unseen tasks and the sensitivity to the type of privileged information for the different methods (see Table \ref{tab:cp2} and Figure \ref{fig:cartpole_task}). Intrinsically, they test the capacity of IMPORT to learn task embeddings.
We reuse the CartPole environment with $\rho=0.05$ but we consider $R$ possible values of $\mu$ (or tasks) at train time, the validation being performed on episodes generated with 100 different values of $\mu$, and the performance being reported over $100$ other values. In this setting, at train time, $\mu$ is encoded as a one-hot vector of size $R$ reflecting which task the agent is currently playing. Note that the informed policy trained on a multi-task CartPole with task index inputs is able to perform optimally. It both allows us to evaluate the ability of the methods to generalize to unseen values of $\mu$, and to compute \emph{task embeddings}. Note that TI and TS models $\mu$ using a multinomial distribution in this setting. Performance is provided in Table \ref{tab:cp2} and in Figure \ref{fig:cartpole_task}. It shows that IMPORT outperforms the other models also in this setting, and that it is thus able to generalize to unseen tasks, even with few tasks at train time. Decreasing performance after $4 \cdot 10^6$ steps can be explained by overfitting on training tasks.

\section{Conclusion}

We have proposed a new policy architecture for learning in multi-task environments. The IMPORT model is trained only on the reward objective, and leverages the informed policy knowledge to discover a good trade-off between exploration and exploitation. It is thus able to learn better strategies than Thompson Sampling approaches, and faster than recurrent neural network policies and Task Inference approaches. Moreover, our approach works well also in non-stationary settings and is able to adapt to generalize to tasks unseen at train time. Learning this model in a  continual learning setting will be investigated in a near future.

\clearpage
\bibliography{bibliography}
\clearpage
\appendix 
\section{Details of the IMPORT algorithm}
The algorithm is described in details in Algorithm \ref{al:import}.

\begin{algorithm*}[h]
        \caption{Details of IMPORT Training}
        \footnotesize
        \begin{algorithmic}
                \STATE Initialize $\sigma, \omega, \theta, \nu$ arbitrarily
                \STATE Hyperparameters: \begin{align*} & \text{ batch size } bs, \text{ L-step return } L, \text{discount factor }\gamma, \text{Adam learning rate } \eta, \text{weighting of the \textbf{(C)} objective }\beta,   \\ &\text{weighting of the \textbf{(B)} objective } \lambda, \text{weighting of the entropy objective } \lambda_h, \text{weighting of the critic objective } \lambda_c\end{align*}
                \STATE $Optim = Adam(\eta)$
     
                \WHILE{training}
                \STATE Collect $(1-\lambda)M$ trajectories according to $\pi_H$ in buffer $B_H$.
                \STATE Collect $\lambda M$ trajectories according to $\pi_{\mu}$ in buffer $B_{\mu}$.
                \STATE $\delta_{\sigma}, \delta_{\omega},\delta_{\theta} = 0, 0 , 0   $
                \STATE
                \STATE  $R^\mu \gets \mathrm{compute\_l\_step\_returns}(B_\mu, L) $ 
                \STATE  $R^H \gets \mathrm{compute\_l\_step\_returns}(B_H, L) $
       
                \STATE
                
                \STATE  $\delta_{\theta, \omega} \mathrel{+}=  \frac{1}{|B_H|} \sum_{b \in B_H} \sum_{t=1}^{T} [ R_t^{\mu, b} - V_{\nu}(s_t^b, z_t^b)] \nabla_{\theta, \omega} \log{\pi_H(a_t^b | s_t^b, z_t^b)} $ 
                \STATE  $\delta_{\theta, \omega} \mathrel{+}=  \frac{\lambda_h}{|B_H|} \sum_{b \in B_H} \sum_{t=1}^{T} \nabla_{\theta, \omega} H\big(\pi_H(a_t^b | s_t^b, z_t^b)\big)$ 
                 \STATE  $\delta_{\omega} \mathrel{-}=  \frac{2\beta}{|B_H|} \sum_{b \in B_H} \sum_{t=1}^{T} [f_H^{\omega}(s_t^b, z_t^b) - f_\mu(s_t^b, \mu_t^b)] \nabla_{\omega} f_H^{\omega}(s_t^b, z_t^b)$
                   \STATE  $\delta_{\nu} \mathrel{-}=  \frac{2 \lambda_c}{|B_H|} \sum_{b \in B_H} \sum_{t=1}^{T} [R_t^{H, b} - V_{\nu}(s_t^b, z_t^b)] \nabla_{\nu} V_{\nu}(s_t^b, z_t^b)$ 
                \STATE
                 \STATE  $\delta_{\theta, \sigma} \mathrel{+}=  \frac{\lambda}{|B_\mu|} \sum_{b \in B_\mu} \sum_{t=1}^{T}  [ R_t^{H, b} - V_{\nu}(s_t^b, \mu_t^b)] \nabla_{\theta, \sigma} \log{\pi_\mu(a_t^b | s_t^b, \mu_t^b)}$
                  \STATE  $\delta_{\theta, \sigma} \mathrel{+}=  \frac{\lambda_h}{|B_\mu|} \sum_{b \in B_\mu} \sum_{t=1}^{T}\nabla_{\theta,\sigma} H\big(\pi_\mu(a_t^b | s_t^b, \mu_t^b)\big)$
                  \STATE  $\delta_{\nu} \mathrel{-}=  \frac{2 \lambda_c}{|B_\mu|} \sum_{b \in B_\mu} \sum_{t=1}^{T} [R_t^{\mu, b} - V_{\nu}(s_t^b, \mu_t^b)] \nabla_{\nu} V_{\nu}(s_t^b, \mu_t^b)$

                \STATE
                \STATE $\theta \gets  Optim(\theta, \delta_{\theta})$
                \STATE $\omega \gets  Optim(\omega, \delta_{\omega})$
                 \STATE $\sigma \gets  Optim(\sigma, \delta_{\sigma})$
                \STATE $\nu \gets  Optim(\nu, \delta_{\nu})$
                \ENDWHILE
        \end{algorithmic}
        \label{al:import}
\end{algorithm*}

\section{Implementation details}

\subsection{Data collection and optimization}
We focus on on-policy training for which we use the actor-critic method A2C \cite{Mnih2016A3C} algorithm. We use a distributed execution to accelerate experience collection. Several worker processes independently collect trajectories. 
As workers progress, a shared replay buffer is filled with trajectories and an optimization step happens when the buffer's capacity $bs$ is reached. After model updates, replay buffer is emptied and the parameters of all workers are updated to guarantee synchronisation.

\subsection{Network architectures}
\label{sub:network_archi}
The architecture of the different methods remains the same in all our experiments, except that the number of hidden units changes across considered environments. A description of the architectures of each method is given in Fig. \ref{fig:architecture}. \\  Unless otherwise specified, MLP blocks represent single linear layers activated with a $tanh$ function and their output size is $hs$.
All methods aggregate the trajectory into an embedding $z_t$ using a GRU with hidden size $hs$. Its input is the concatenation of representations of the last action $a_{t-1}$ and current state $s_t$ obtained separately.  For bandits environments, the current state corresponds to the previous reward. 
TS uses the same GRU architecture to aggregate the history into $z_t$. 

All methods use a $softmax$ activation to obtain a probability distribution over actions. \\
The use of the hidden-state $z_t$ differs across methods.
While \textbf{RNNs} only use $z_t$ as an input to the policy and critic, both \textbf{TS} and \textbf{TI} map $z_t$ to a belief distribution that is problem-specific, e.g. Gaussian for control problems, Beta distribution for bandits, and a multinomial distribution for Maze and CartPole-task environments. For instance, $z_t$ is mapped to a Gaussian distribution by using two MLPs whose outputs of size $|\mu|$ correspond to the mean and variance. The variance values are mapped to $[0,1]$ using a $sigmoid$ activation.\\
\textbf{IMPORT} maps $z_t$ to an embedding $f_H$, whereas the task embedding 
$f_\mu$ is obtained by using a $tanh$-activated linear mapping of $\mu_t$. Both embeddings have size $hs_{\mu}$, tuned by cross-validation onto a set of validation tasks. The input of the shared policy head $\phi$ is the embedding associated with the policy to use, i.e. either  $f_H$ when using $\pi_H$ or $f_\mu$ when using $f_\mu$.
For the Maze3d experiment, the pixel input $s_t$ is fed into three convolutional layers (with output channels 32) and LeakyReLU activation (kernel size are respectively 5, 5 and 4 and stride is 2). The output is flattened, linearly mapped to a vector of size $hs$ and $tanh$-activated.

\subsection{Results preprocessing}
\label{ap:results_preprocessing}
We run each method for different hyperparameter configurations, specified in Appendix \ref{ap:experiments}, and choose the best hyperparameters using grid-search. 
We separate task sets into disjoints training, validation and testing sets. During training, every 10 model updates, the validation performance is measured by running on 100 episodes with $\mu$ taken from the validation tasks. Similarly, the test performance is measured using $100$ testing tasks.

Each pair (method, set of hyperparameters) was trained with 3 seeds. 
For each method, we define the best set of hyperparameters as follows.
First, for each seed, find the best validation performance achieved by the model over the course of training. The score of a set of hyperparameters is then the average of this performance over seeds.  The best set of hyperparameters is the one with maximum score.

\paragraph{Plots.} 
Each curve was obtained by averaging over 3 seeds already-smoothed test performance curves. The error bars correspond to standard deviations. Smoothing is done with a sliding window of size $11$. For each method, we only plot the method with the best set of hyperparameters, as defined above. The x-axes of the plots correspond to environment steps.
 
\paragraph{Tables.}  Results in Tables \ref{tab:bandit}, \ref{tab:cp}, \ref{tab:acro} and \ref{tab:cp2} correspond to the mean and standard deviation (over seeds) of the test performance obtained by the policy extracted from the model with the best set of hyperparameters at maximum validation performance.

\section{Experiments}
\label{ap:experiments}
In this section, we explain in deeper details the environments and the set of hyper-parameters we considered. We add learning curves of all experiments to supplement results from Table  \ref{tab:bandit}, \ref{tab:cp}, \ref{tab:acro} and \ref{tab:cp2} in order to study sample efficiency.
Note that for all experiments but bandits, $\mu$ is normalized to be in $[-1,1]^D$ where $D$ is the task descriptor dimension.

Hyperparameters ranges specified in Table \ref{table:hp_common} are kept constant on all environments. Environment-specific hyperparameters (hidden size $hs$, belief distribution for TS/TI, ...) will be specified in Appendix \ref{ap:experiments}.

\begin{table}[h]
 \centering
    \begin{tabular}{cc}\toprule
        Hyperparameter & Considered values \\ \toprule
        $bs$ & 4  \\
        $\gamma$ & 0.95 \\
        clip gradient & 40 \\
        $\eta$ & $\{1e^{-3}, 3e^{-3}\}$ \\
        $\lambda_h$ & $\{1e^{-1}, 1e^{-2}, 1e^{-3}\}$  \\
        $\lambda_c$ & $\{1, 1e^{-1}, 1e^{-2}\}$ \\
        $\beta$ & $\{1, 1e^{-1}, 1e^{-2}, 1e^{-3}\}$  \\
        $\lambda$ & $\{0.5, 0.75\}$  \\
        \bottomrule
    \end{tabular}
    \caption{Hyperparameters range}
    \label{table:hp_common}
\end{table}

\subsection{Bandits}
\label{bandit}
     
At every step, the agent pulls one of K arms, and obtains a stochastic reward drawn from a Bernoulli distribution with success probability $\mu_i$, where $i$ is the arm id. The goal of the agent is to maximize the cumulative reward collected over $100$ steps. At test time, the agent does not know  $\mu = (\mu_1,..., \mu_K)$ and only observes the reward of the selected arm.

$\mu$ is sampled according to the following multivariate random variable with constants $\mu_{max}$ and $\mu_{min}$ fixed beforehand:
\begin{itemize}
    \item an optimal arm $i^*$ is sampled at random in $\mathcal{U}([\![1,K ]\!])$ and $\mu_{i^*}=\mu_{max}$
    \item $\forall i \neq i^*,  \mu_{i} \sim U([0,\mu_{min}])$
\end{itemize}
At each time-step, there is a probability $\rho$ to sample a new value of $\mu$.

We consider different configurations of this generic schema with $\rho = 0.05, \ K \in \{5, 10, 20\}, \ \mu_{max} \in \{0.5, 0.9\}, \ \mu_{min} = 0.1$. 

All methods use $hs=hs_{\mu}=32$ and the belief distribution is either a Beta distribution or Gaussian. Other hyperparameters are presented in Table \ref{table:hp_common}.

\begin{figure*}[t]
\centering
\includegraphics[width=0.48\linewidth]{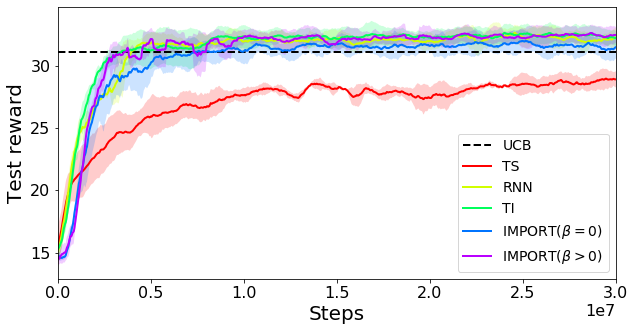}
\includegraphics[width=0.48\linewidth]{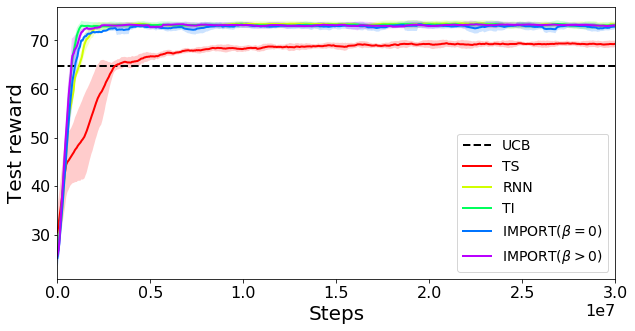}
\caption{Training curves on the bandits environment with $K=5$ and respectively $\mu_{min}=0.5$ (left) and $\mu_{min}=0.9$ (right).}
\label{fig:bandits5}
\end{figure*}

\begin{figure*}[t]
\centering
\includegraphics[width=0.48\linewidth]{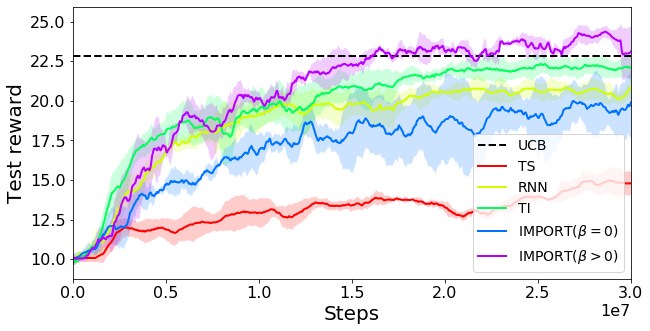}
\includegraphics[width=0.48\linewidth]{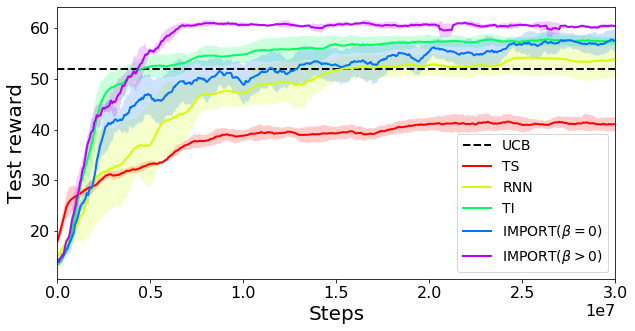}
\caption{Training curves on the bandits environment with $K=10$ and respectively $\mu_{min}=0.5$ (left) and $\mu_{min}=0.9$ (right).}
\label{fig:bandits10}
\end{figure*}

\begin{figure*}[t]
\centering
\includegraphics[width=0.48\linewidth]{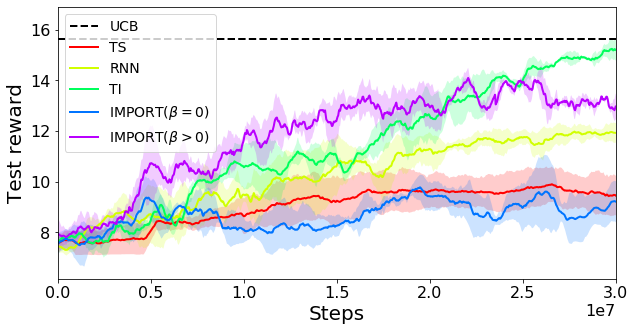}
\includegraphics[width=0.48\linewidth]{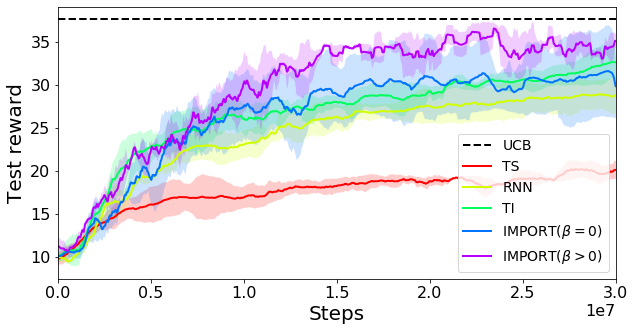}
\caption{Training curves on the bandits environment with $K=20$ and respectively $\mu_{min}=0.5$ (left) and $\mu_{min}=0.9$ (right).}
\label{fig:bandits20}
\end{figure*}

Since the setting with $K=5$ is fairly easy to solve, RNN, TI and IMPORT perform on par (see Fig. \ref{fig:bandits5}). TS performs worse as it is sub-optimal in non-stationary environments. For $K=10$ (Fig. \ref{fig:bandits10}), IMPORT largely outperforms other methods. When $\mu_{min}=0.5$, the gap between the optimal arm and the second best can be small. The optimal policy does not necessarily stick to the best arm and learning is slower. When $K=20$ (Fig. \ref{fig:bandits20}), learning is harder and the UCB baseline is better.

\subsection{CartPole.}
\label{ap:cartpole}

We consider the classic CartPole control environment where the environment dynamics change within a set $\mathcal{M}$ ($|\mu|=5$) described by the following physical variables: gravity, cart mass, pole mass, pole length, magnetic force.
Their respective  pre-normalized domains are $[4.8,14.8], [0.5, 1.5], [0.01, 0.19], [0.2, 0.8], [-10, 10]$. 
Knowing some components of $\mu$ might not be required to behave optimally. The discrete action space is $\{-1,1\}$.

$\mu$'s are re-sampled at each step with probability $\rho \in \{0, 0.05, 0.1\}$. Episode length is $T=100$.

All methods use $hs=hs_{\mu}=16$ and the belief distribution is Gaussian. Other hyperparameters are presented in Table \ref{table:hp_common}.

\begin{figure*}[t]
\includegraphics[width=0.48\linewidth]{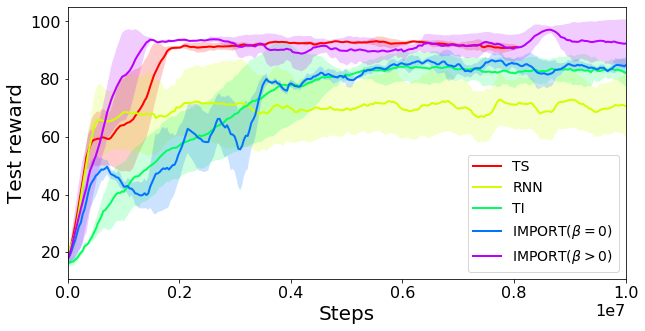}
\includegraphics[width=0.48\linewidth]{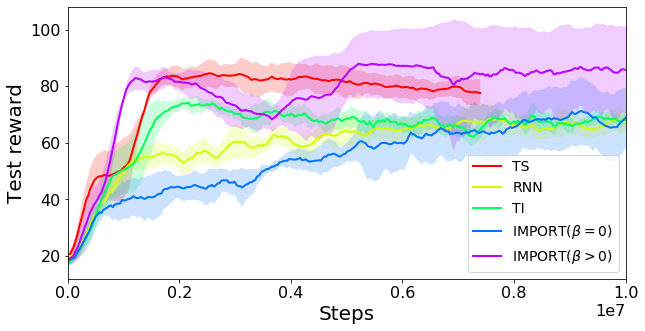}
\includegraphics[width=0.48\linewidth]{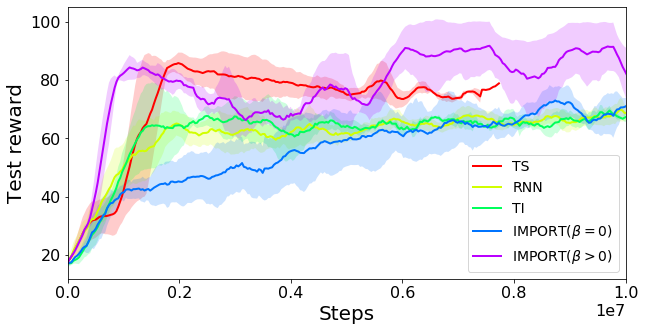}\hfill
\caption{Training curves on the CartPole environment with respectively $\rho = 0$ (top left), $\rho = 0.05$ (top right), $\rho = 0.1$ (bottom)}.
\label{fig:cartpolecurves}
\end{figure*}

Figure \ref{fig:cartpolecurves} shows IMPORT's performance and sample efficiency is greatly superior to other methods. IMPORT($\beta>0$) performs on par or worse than TI, which proves that IMPORT main advantage is the auxiliary supervised loss. TI performs dramatically worse, showing reconstructing the entire $\mu$ is not optimal.

\subsection{CartPole-task}
\label{ap:cartpole_task}

To study how the different methods deal with cases where no meaningful physical parameters of the system is available, as well as studying their performance on tasks that were not seen during training, we conduct a new set of experiments in the CartPole environment described below. In this new set of experiments, $\mu$ represents the task identifier of the considered $\mu$-MDP. Here $\mu$ is a one-hot encoding of the MDP, thus containing no relevant information on the world dynamics. 
To assess generalization on unseen tasks, we consider a training task set of $R$ different tasks where the underlying dynamics parameters are sampled in the same way than for the usual CartPole environment.Validation and testing task sets are then additional disjoints set of $100$ tasks (thus,  there is no overlap between train, validation and test task sets).

$\mu$'s are re-sampled at each step with probability $\rho \in \{0, 0.1\}$. Episode length is $T=100$.

Considered hyperparameters in CartPole-task are the same than the ones in CartPole except the belief distribution is multinomial.

\begin{figure*}[t]
\centering
\includegraphics[width=0.48\linewidth]{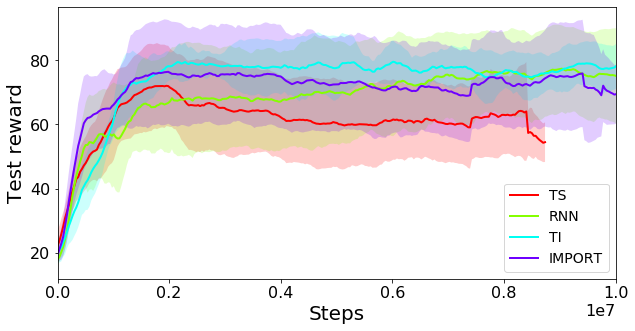}
\includegraphics[width=0.48\linewidth]{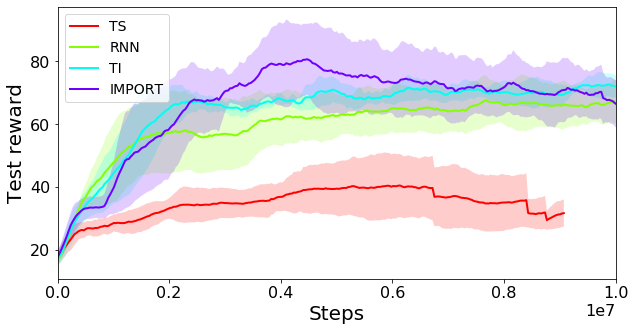}
\caption{Training curves on the CartPole-task environment with $R=10$ and respectively $\rho = 0$ (left), $\rho = 0.1$ (right)}.
\label{fig:cartpoletask10curves}
\end{figure*}

\begin{figure*}[t]
\centering
\includegraphics[width=0.48\linewidth]{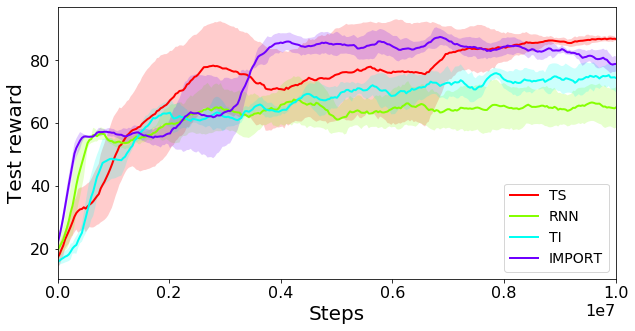}
\includegraphics[width=0.48\linewidth]{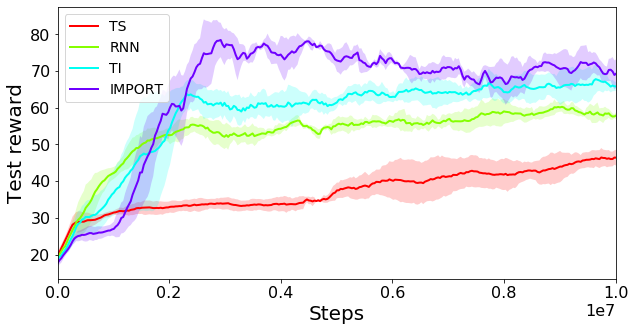}
\caption{Training curves on the CartPole-task environment with $R=100$ and respectively$\rho = 0$ (left), $\rho = 0.1$ (right)}.
\label{fig:cartpoletask100curves}
\end{figure*}

In stationary environments, all methods are roughly equivalent in performance (Figures \ref{fig:cartpoletask10curves}, \ref{fig:cartpoletask100curves}). Indeed, in control problems, there is no need of a strong exploration policy since the underlying physics can be inferred from few transitions. 
When the environment is non-stationary, IMPORT is significantly better than the baselines. In the end,  these experiments suggest that,  in the stationary setting, all methods are able to generalize to unseen tasks on that environment. In the non-stationary setting however, IMPORT significantly outperforms the baselines.

\subsection{Acrobot}
\label{ap:acrobot}

Acrobot consists of two joints and two links, where the joint between the two links is actuated. Initially, the links are hanging downwards, and the goal is to swing the end of the lower link up to a given height. Environment dynamics  are determined by the length of the two links, their masses, their maximum velocity.
Their respective pre-normalized domains are $[0.5,1.5], [0.5, 1.5], [0.5, 1.5],[0.5, 1.5], [3 \pi, 5\pi], [7\pi, 11\pi]$. 
Unlike CartPole, the environment is stochastic because the simulator applies noise to the applied force. The action space is $\{-1,0,1\}$. 
We also add an extra dynamics parameter which controls whether the action order is inverted, i.e. $\{1,0,-1\}$, thus $|\mu|=7$. 

$\mu$'s are re-sampled at each step with probability $\rho \in \{0, 0.01, 0.05\}$. Episode length is $500$.

All methods use $hs=hs_{\mu}=64$ and the belief distribution is Gaussian. Other hyperparameters are presented in Table \ref{table:hp_common}.

\begin{figure*}[t]
\includegraphics[width=0.48\linewidth]{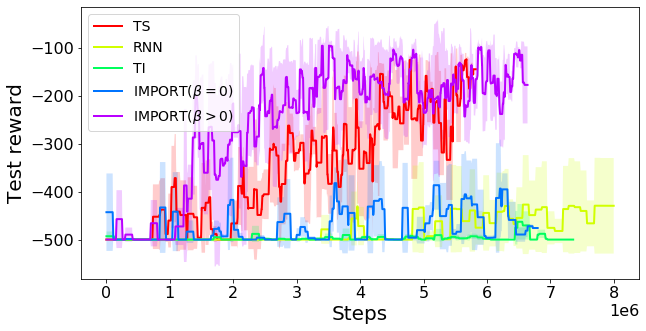}
\includegraphics[width=0.48\linewidth]{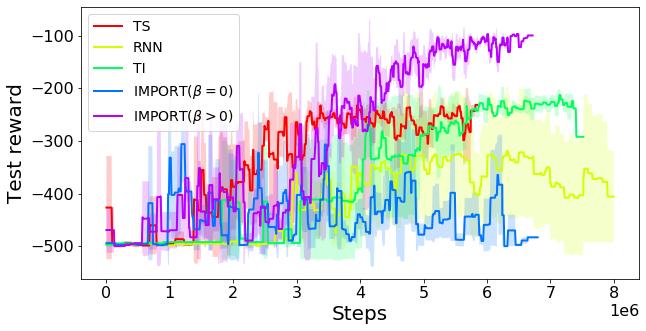}
\includegraphics[width=0.48\linewidth]{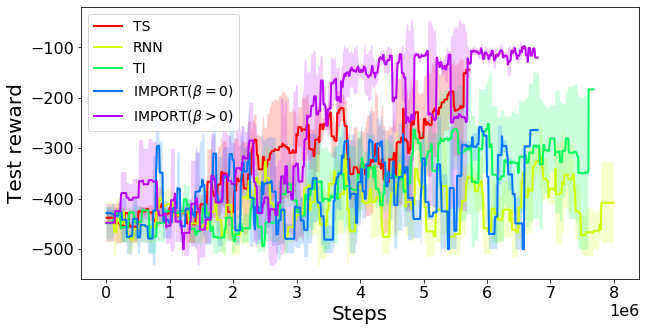}
\caption{Training curves on the Acrobot environment respectively for $\rho = 0$ (top left), $\rho = 0.01$ (top right), $\rho = 0.05$ (bottom)}.
\label{fig:acrobot}
\end{figure*}

IMPORT outperforms all baselines in every settings (Fig. \ref{fig:acrobot}).

 \subsection{Maze environments}
\label{ap:maze}
Maze2d/Maze3d are grid-world environments with two possible goals positioned at two different locations and a sign that indicates which goal is activated when visited. The value of $\mu=+1$ or $-1$ denotes which of the two goals is active. $\mu$ is sampled at random at the beginning of each episode. The agent can access this information by moving to a \textit{sign} location. It has to reach the goal in at most $100$ steps -- the optimal policy is able to achieve this objective in 19 steps -- to receive a reward of $+1$ and the episode stops. If the agent reaches the wrong goal, it receives a reward of $-1$ and the episode stops. The agent has three actions (forward, turn left, turn right) and the environment has been implemented using \emph{MiniWorld}~\citep{gym_miniworld}.
 
 We propose two versions of the same grid-world environment but with different inputs given to the agent. The Maze2d version where the agent observes its absolute coordinates (and eventually the goal location when going over the \emph{sign}, otherwise a placeholder s.t. $0$). The Maze3d version where the agent observes a highly-dimensional ($600 \times 400$) image, the sign information is encoded as a fourth dimension (either zeros or $\mu$ values when the sign is touched) concatenated to the RGB image. Note that it is an environment with a sparse reward (sionce there is no reward when reaching the sign) that is very complex to solve because the policy has to learn to discover the sign location, to associate the sign information with the sign location, to memorize the \emph{sign} information, and to reach the goal. In this setting, TS and TI are using a \textit{Bernoulli} distribution to model $\mu$.

In both cases, the maze's width and length are 12 with coordinates going from $-6$ to $6$ in both directions. The goal locations are $(-5, 5)$ and $(5,5)$.
In order to adjust the difficulty of solving the environment, we tried two scenarios: 
\begin{itemize}
    \item Scenario A: The sign is located on $(0,0)$ and the agent starts in position $(0,-5)$. The agent does not waste time going to the sign as it is on its road.
    \item Scenario B: The sign is located on $(0,-5)$ and the agent starts in position $(0,0)$. This requires the agent to go to the bottom of the maze first, then remember the goal location, and finally go to the activated goal. This is a very hard exploration problem.
\end{itemize}

In the main article, results are reported on scenario A with a single seed. We report here complete results on the two scenarios on multiple seeds.

All methods use $hs \in \{16,32\}$, $hs_{\mu} \in \{4, 8\}$ and the belief distribution is Bernoulli. Other hyperparameters are presented in Table \ref{table:hp_common}.

\begin{figure*}[t]
\centering
\includegraphics[width=0.48\linewidth]{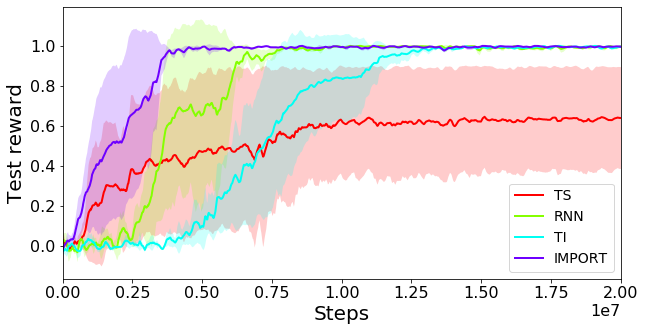}
\includegraphics[width=0.48\linewidth]{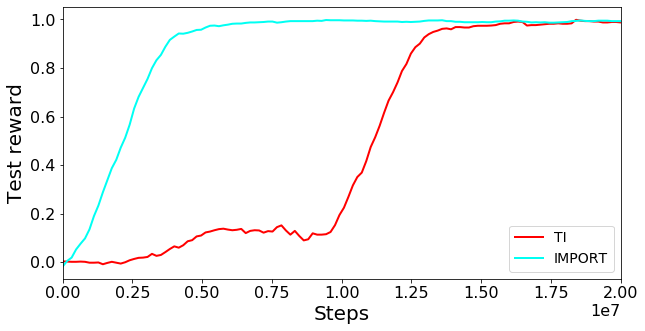}
\caption{Training curves on the scenario A of Maze2d (top) and Maze3d (bottom)}
\label{fig:mazeA}
\end{figure*}

\begin{figure*}[t]
\centering
\includegraphics[width=0.48\linewidth]{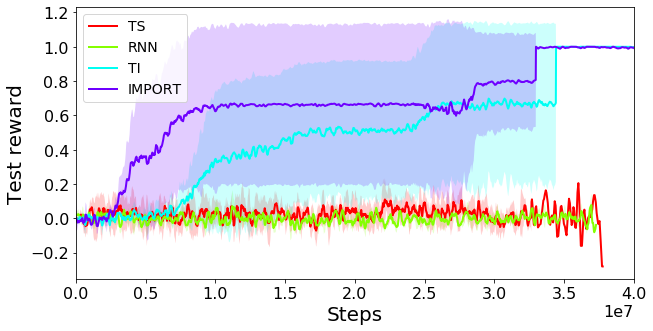}
\caption{Training curves on the scenario B of Maze2d. We were not able to have this scenario solved with image inputs by the different methods }
\label{fig:mazeB}
\end{figure*}

IMPORT outperforms other methods on Scenario A in both Maze2d and Maze3d (Fig. \ref{fig:mazeA}) in sample efficiency. Due to time constraints, we only ran Maze3d on just one seed. In Scenario B (Fig. \ref{fig:mazeB}), IMPORT is a bit more sample efficient in Maze2d. We were not able to have scenario B solved with image inputs by the different methods. 

\clearpage

\bibliographystyle{icml2020}
\end{document}